\documentclass{article}
\usepackage{ijcai26}

\usepackage{times}
\usepackage{soul}
\usepackage{url}
\usepackage[hidelinks]{hyperref}
\usepackage[utf8]{inputenc}
\usepackage[small]{caption}
\usepackage{graphicx}
\usepackage{amsmath}
\usepackage{amsthm}
\usepackage{booktabs}
\usepackage{algorithm}
\usepackage[switch]{lineno}

\usepackage{booktabs}       
\usepackage{amsfonts}       
\usepackage{nicefrac}       
\usepackage{microtype}      
\usepackage{xcolor}         
\usepackage{graphicx}
\usepackage{float}
\usepackage{amsmath}
\usepackage{multirow}
\usepackage{algpseudocode}

\algnewcommand\Input{\item[\textbf{Input:}]}
\algnewcommand\Output{\item[\textbf{Output:}]}

\title{Property Prediction of Stacked Bilayer Materials: A Multimodal Learning Approach}

\author{
An Vuong$^1$
\and
Minh-Hao Van$^1$\and
Chen Zhao$^{2}$\And
Xintao Wu$^1$ \\
\affiliations
$^1$University of Arkansas\\
$^2$Baylor University\\
\emails
\{anv, haovan\}@uark.edu,
chen\_zhao@baylor.edu,
xintaowu@uark.edu
}

\begin{document}

\maketitle

\begin{abstract}
    AI for materials science is a critical topic within AI for science, aiming to accelerate materials discovery and produce accurate property predictions. Bilayer 2D material stacking is essential for exploring new materials with novel functions and inherent phenomena, enabling the creation of new 2D bilayers for diverse real-world applications. Research on bilayer vdWs materials has made significant progress from experimental and computational perspectives.  Various bilayer materials have been successfully synthesized experimentally and the increasing utilization of high-throughput computing technology has constructed several computational two-dimensional materials databases. However, the use of AI to model bilayer stacking and predict new properties remains underexplored, necessitating further research studies. In this work, we propose a novel multimodal learning approach to study the interfaces between dissimilar materials that jointly enable new or multiple functions, and to predict new properties arising from the vertical integration (stacking) of different functional material layers under given configurations. Comprehensive experiments demonstrate the effectiveness and efficiency of our approach compared to baseline methods. Our code is available at \url{https://github.com/AnVuong123/bimat_ml}. 
  
\end{abstract}

\section{Introduction}

Bilayer materials are two-dimensional structures composed of two individual layers of 2D materials that are stacked on top of each other. The layers are held together by weak van der Waals forces, but the way they are stacked can be manipulated, a process that allows for fine-tuning of the material's properties. Stacking layers can create new electronic, optical, and mechanical properties that are not present in the single-layer (monolayer) versions of the materials. This is especially true when the layers are twisted relative to each other, creating a moiré superlattice.  For example, twisted bilayers (TBL) of 2D materials have emerged as a versatile platform for novel interface phenomena \cite{li2010observation}. The two layers that form these twisted structures when isolated do not exhibit ferroelectricity, indicating that the effect arises from interlayer coupling, charge redistribution, and moiré-induced asymmetry rather than lattice displacements.

There has been increasing necessity of considering various stacking modes in 2D vdWs bilayer structures, including patterns and sequences, which significantly affect the material properties. A key task is to predict the properties of stacked bilayer materials from their stacking configuration. Computational simulation such as density functional theory (DFT) has been a popular method for calculating the intrinsic properties of materials on the basis of electron density. When stacking 2D materials, the high degree of freedom of stacking results in a large material space.  The high computational cost of DFT makes it difficult to study stacked 2D materials.
Graph neural networks (GNNs) have been introduced into materials science for processing molecular/crystal data with non-Euclidean structures and many works \cite{xie2018crystal,choudhary2021atomistic} demonstrate high performance  because they effectively represent and capture graph structures from  crystallographic information files (CIFs) of materials.  However, for stacked two dimensional materials, inter-layer atomic interactions are  bound by weak van der Waals forces and intra-layer atomic interactions are bound by chemical bounds. Directly applying GNN models on the stacked materials cannot differentiate between intralayer and interlayer interactions, thus being unable to achieve accurate property predictions of the properties of stacked 2D materials. 

In this paper, we propose a unified multimodal learning framework for modeling bilayer materials, namely BiMat-ML, that jointly models monolayer material structures, stacking configurations, and intrinsic monolayer properties to predict target properties of stacked bilayer materials. By explicitly integrating information across these complementary modalities, our approach captures both intra-layer chemistry and inter-layer stacking effects within a single predictive model. Our proposed framework is model-agnostic,  applicable to diverse stacking configurations of both homobilayer and heterobilayer settings. Experimental results demonstrate that our method achieves prediction accuracy comparable to density functional theory (DFT) calculations while providing orders-of-magnitude improvements in computational efficiency. These results highlight the potential of multimodal learning as a general and scalable paradigm for rapid screening and inverse design of stacked 2D materials.

\section{Related Work}
 
Bilayer materials are atomically thin structures consisting of two stacked layers of 2D materials, such as graphene or transition metal dichalcogenides, held together by van der Waals (vdWs) forces. These stacked layers have unique electronic, optical, and mechanical properties that can be tuned by controlling the stacking pattern between the layers.
The different stacking patterns can be created through rotations around the vertical axis or horizontal layer sliding \cite{wang2021stacking}. 
Research on bilayer vdWs materials has made significant progress from experimental and computational perspectives.  Various bilayer materials have been successfully synthesized experimentally and the increasing utilization of high-throughput computing technology has constructed several computational two-dimensional materials databases such as C2DB \cite{gjerding2021recent}, MC2D \cite{mounet2018two}, 2DMatPedia \cite{zhou20192dmatpedia}, BMDS \cite{barik2023high}, BiDB \cite{pakdel2024high}, and HetDB \cite{sauer2025dispersion}.  These databases provide a rich theoretical foundation for materials research and significantly accelerate the research and development of 2D materials.  
To simply the high-throughtput computational process, the authors in \cite{zhang2025pyhtstack2d} developed a specialized Python package, PyHTStack2D, designed for the efficient High-Throughput Stacking of 2D materials, including both homo- and hetero- structures. The package assists in generating input files and shell scripts for batch computation submissions to the Vienna Ab initio Simulation Package (VASP).

Advanced AI models are catalyzing a transformative shift in materials science.  Many AI algorithms have been developed to support various discovery tasks such as  atomistic simulation, property prediction, materials structure design and discovery, process planning and optimization \cite{van2025survey,merchant2023scaling}. Unlike 3D materials, the AI development for bilayer 2D materials received very little study.  
\cite{chen2024structural} developed a structural embedding method for property prediction of stacked bilayer materials.  The developed structure-embedded PAINN (SE-PAINN) independently considers intra-layer and inter-layer neighbors when determining the connectivity between nodes (atoms) and uses two symmetric neural networks to handle intra-layer interactions and inter-layer interactions separately. They showed SE-PAINN can approximately reproduce the DFT calculation results of predicting the binding energy and band gap of twisted stacked 2D materials with significantly reduced computational cost. However, this method still requires the access of CIFs of stacked bilayer materials in both training and inference, thus being unable to support the bilayer property prediction task when CIFs of only monolayer materials are available. 

\section{Property Prediction of Stacked Bilayer Materials via Multimodal Learning}

Bilayer materials are created by stacking two monolayers according to a specified configuration. In typical bilayer construction workflows, one monolayer is assigned as the bottom layer and the other as the top layer. Each monolayer is described by a crystallographic information file (CIF), which stores its crystal structure data. The objective of this work is to predict the target properties of the resulting bilayer material using the stacking configuration, the CIF files and the known properties of individual monolayers.
In this section, we first present our framework BiMat-ML and then describe in detail each component including the graph encoder used for learning representation of each monolayer, and stacking configuration autoencoder used for learning configuration representation.

\subsection{Framework}

\begin{figure*}[ht]
    \centering
    \includegraphics[width=.90\linewidth]{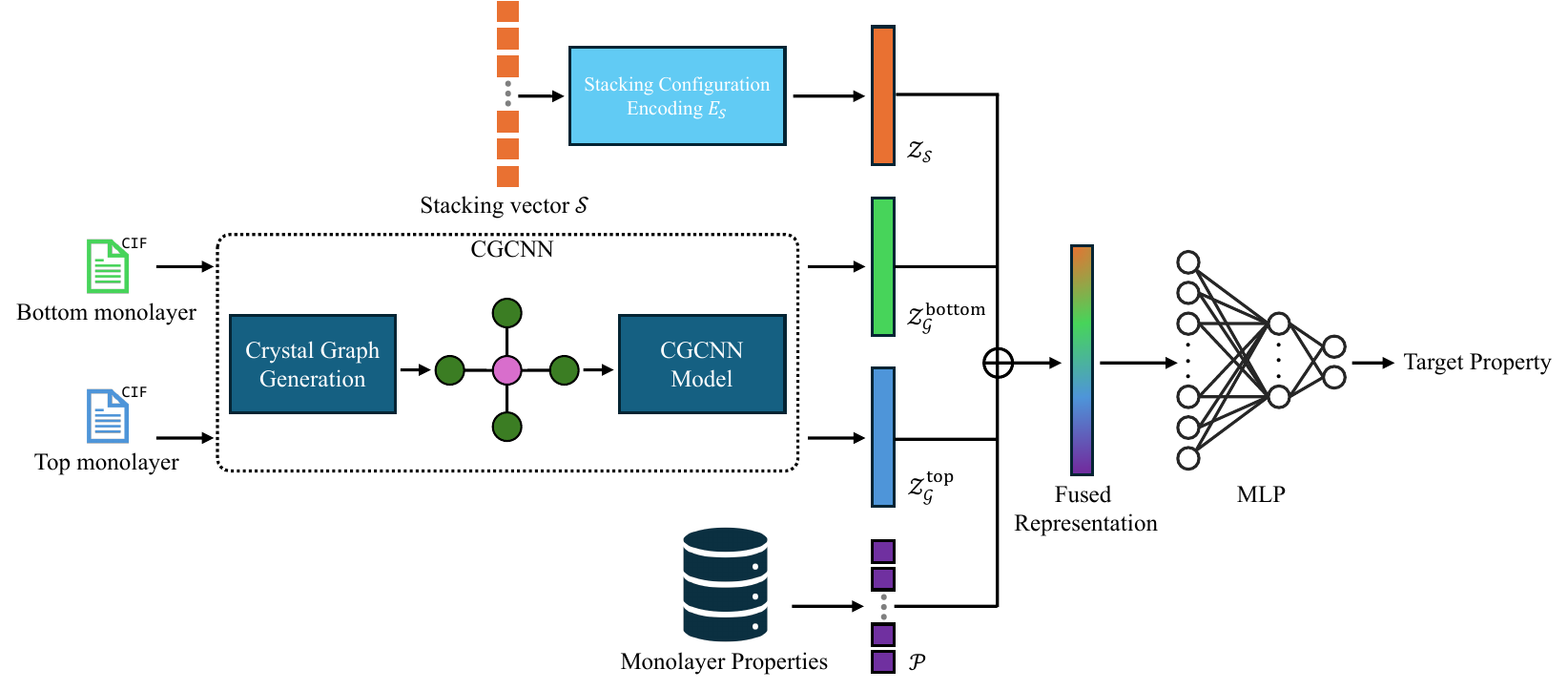}
    \caption{Architecture of property prediction of stacked bilayer material via multimodal learning (BiMat-ML)}
    \label{fig:architecture}
\end{figure*}

We propose a unified multimodal learning framework that jointly models monolayer material structures, stacking configuration, and properties of monolayer materials, to predict target properties of bilayer materials.
Formally, the bilayer material dataset is represented as
$\mathcal{D} = \{ (\mathcal{G}^{\mathrm{bottom}}, \mathcal{P}^{\mathrm{bottom}}, \mathcal{G}^{\mathrm{top}}, \mathcal{P}^{\mathrm{top}}, \mathcal{S}), \mathcal{Y} \}$,
where $\mathcal{G}^{\mathrm{bottom}}$ ($\mathcal{G}^{\mathrm{top}}$) denotes the crystal graph structure of the bottom (top) monolayer,
$\mathcal{P}^{\mathrm{bottom}}$ ($\mathcal{P}^{\mathrm{top}}$) represents known properties of the bottom (top) monolayer,
$\mathcal{S}$ denotes the stacking configuration information,
and $\mathcal{Y}$ is the target property of the bilayer material.
The objective is to learn a mapping function
\begin{equation}
f_{\theta} : (\mathcal{G}^{\mathrm{bottom}}, \mathcal{P}^{\mathrm{bottom}}, \mathcal{G}^{\mathrm{top}}, \mathcal{P}^{\mathrm{top}}, \mathcal{S}) \rightarrow \mathcal{Y},
\end{equation}
where $\theta$ denotes the set of learnable parameters.
Figure~\ref{fig:architecture} illustrates the architecture of BiMat-ML that consists of three main components:
(i) a graph encoder $E_{\mathcal{G}}$ that encodes monolayer crystal graphs into graph representations,
(ii) a stacking configuration encoder $E_{\mathcal{S}}$ that encodes stacking configurations into latent representations, and
(iii) a multimodal fusion module that integrates structural, stacking, and property information into a joint embedding for property prediction.

\begin{algorithm}[h]
\caption{Property Prediction of Stacked Bilayer Material via Multimodal Learning (BiMat-ML) }
\label{alg:multimodal_fusion}
\begin{algorithmic}[1]
\setlength{\baselineskip}{1.2\baselineskip}
\Input Bottom and top monolayer CIF $\mathcal{C}^{\mathrm{bottom}}, \mathcal{C}^{\mathrm{top}}$;
       stacking configurations $\mathcal{S}$;
       monolayer properties $\mathcal{P}^{\mathrm{bottom}}, \mathcal{P}^{\mathrm{top}}$
\Output Predicted bilayer property $\hat{\mathcal{Y}}$

\State $\mathcal{Z}^{\mathrm{bottom}}_{\mathcal{G}} \gets E_{\mathcal{G}}(\mathcal{C}^{\mathrm{bottom}})$
\State $\mathcal{Z}^{\mathrm{top}}_{\mathcal{G}} \gets E_{\mathcal{G}}(\mathcal{C}^{\mathrm{top}})$

\State $\mathcal{Z}_{\mathcal{S}} \gets E_{\mathcal{S}}(\mathcal{S})$

\State $\mathcal{Z} \gets
\mathcal{Z}^{\mathrm{bottom}}_{\mathcal{G}}
\oplus
\mathcal{P}^{\mathrm{bottom}}
\oplus
\mathcal{Z}^{\mathrm{top}}_{\mathcal{G}}
\oplus
\mathcal{P}^{\mathrm{top}}
\oplus
\mathcal{Z}_{\mathcal{S}}
$

\State $\hat{\mathcal{Y}} \gets \mathrm{MLP}(\mathcal{Z})$
\State $\mathcal{L} \gets | \hat{\mathcal{Y}} - \mathcal{Y} |$
\State Update all parameters by minimizing $\mathcal{L}$
\State \Return $\hat{\mathcal{Y}}$
\end{algorithmic}
\end{algorithm}

Algorithm~\ref{alg:multimodal_fusion} shows the pseudo code of BiMat-ML training algorithm. 
Specifically, we use CGCNN encoder (denoted as $E_\mathcal{G}$) to learn graph-level embeddings of each monolayer. $E_\mathcal{G}$ first constructs monolayer crystal graphs ($\mathcal{G}^{\mathrm{bottom}}$,$\mathcal{G}^{\mathrm{top}}$ ) from monolayer CIFs ($\mathcal{C}^{\mathrm{bottom}}$, $\mathcal{C}^{\mathrm{top}}$) and then  maps crystal graphs to graph-level embeddings ($\mathcal{Z}^{\mathrm{bottom}}_{\mathcal{G}}$, $\mathcal{Z}^{\mathrm{top}}_{\mathcal{G}}$). We than use the stacking configuration encoder $E_{\mathcal{S}}$ to map stacking configurations $\mathcal{S}$ to stacking embeddings.
These representations, together with the monolayer material properties $\mathcal{P}$, are subsequently fused
to form a unified multimodal representation.
As the learned graph representations $\mathcal{Z}^{\mathrm{bottom}}_{\mathcal{G}}, \mathcal{Z}^{\mathrm{top}}_{\mathcal{G}} \in \mathbb{R}^{d_{\mathcal{G}}}$,   stacking configuration representation $\mathcal{Z}_{\mathcal{S}} \in \mathbb{R}^{d_{\mathcal{S}}}$, and monolayer property  $\mathcal{P}^{\mathrm{bottom}}, \mathcal{P}^{\mathrm{top}} \in \mathbb{R}^{d_{\mathcal{P}}}$, all modality-specific representations are concatenated to form a unified embedding
\[
\mathcal{Z}
=
\mathcal{Z}^{\mathrm{bottom}}_{\mathcal{G}}
\oplus
\mathcal{P}^{\mathrm{bottom}}
\oplus
\mathcal{Z}^{\mathrm{top}}_{\mathcal{G}}
\oplus
\mathcal{P}^{\mathrm{top}}
\oplus
\mathcal{Z}_{\mathcal{S}}
\in
\mathbb{R}^{2(d_{\mathcal{G}} +d_{\mathcal{P}}) + d_{\mathcal{S}}}.
\]

The joint representation $\mathcal{Z}$ is subsequently passed through a multi-layer perceptron (MLP) to predict the target bilayer material property.
The entire framework is trained end-to-end by minimizing the mean absolute error loss
$\mathcal{L} = | \hat{\mathcal{Y}} - \mathcal{Y} |$.

\subsection{Monolayer Representation via Graph Encoder}

\begin{algorithm}[h]
\caption{CGCNN-based Graph Encoding ($E_{\mathcal{G}}$)}
\label{alg:cgcnn_encoder}
\begin{algorithmic}[1]
\setlength{\baselineskip}{1.2\baselineskip}
\Input CIF file $\mathcal{C}$; number of graph convolution layers $T$
\Output Graph representation $\mathcal{Z}_{\mathcal{G}}$

\State Parse $\mathcal{C}$ to obtain atomic species and coordinates.
\State For each atom $i$, identify neighboring atoms within cutoff radius $R$ and form a local edge set $\mathcal{E}_i$.
\State Construct crystal graph $\mathcal{G}=(\mathcal{V},\mathcal{E})$.
\State Initialize node features $\{v_i^{(0)} \mid i \in \mathcal{V}\}$.
\State Initialize edge features $\{u_{(i,j)_k} \mid (i,j)_k \in \mathcal{E}_i\}$.

\For{$t=0$ \textbf{to} $T-1$}
  \For{each atom $i \in \mathcal{V}$}
    \State $v_i^{(t+1)} \gets v_i^{(t)}$
    \For{each $(i,j)_k \in \mathcal{E}_i$}
      \State $z_{(i,j)_k}^{(t)} \gets v_i^{(t)} \oplus v_j^{(t)} \oplus u_{(i,j)_k}$
      \State 
      $\begin{aligned}[t]
        v_i^{(t+1)} \gets & v_i^{(t+1)} +
      \sigma(z_{(i,j)_k}^{(t)} W_f^{(t)} + b_f^{(t)})
      \odot\\
      &g(z_{(i,j)_k}^{(t)} W_s^{(t)} + b_s^{(t)})
        \end{aligned}$
    \EndFor
  \EndFor
\EndFor

\State $\mathcal{Z}_{\mathcal{G}} \gets \frac{1}{|\mathcal{V}|}\sum_{i} v_i^{(T)}$
\State \Return $\mathcal{Z}_{\mathcal{G}}$
\end{algorithmic}
\end{algorithm}

Algorithm~\ref{alg:cgcnn_encoder} describes steps to  use  CGCNN to map a crystal structure to a fixed-dimensional representation. The structure information in a CIF file $\mathcal{C}$ is first parsed to extract atomic species, lattice vectors, and atomic coordinates. Based on the atomic coordinates and periodic boundary conditions, interatomic distances are computed. For each atom $i$, neighboring atoms are identified based on interatomic distances within a cutoff radius $R$, and up to a maximum of $N$ nearest neighbors are retained to form a local edge set $\mathcal{E}_i$.
The crystal graph is represented as $\mathcal{G}=(\mathcal{V},\mathcal{E})$ where $\mathcal{V}$ is the set of atoms and $\mathcal{E} = \bigcup_{i=1}^{|\mathcal{V}|}\mathcal{E}_i$. Each atom $i \in \mathcal{V}$ is initialized with an elemental feature vector $v_i^{(0)}$, obtained by mapping its atomic identity to a fixed-length embedding following the CGCNN framework. Due to periodic boundary conditions, multiple edges $(i,j)_k$ may exist between the same atom pair, where $(i,j)_k$ denotes the $k$-th bond connection between atom $i$ and atom $j$. For each edge $(i,j)_k$, the corresponding interatomic distance $d_{(i,j)_k}$ is encoded into a fixed-dimensional edge feature vector $u_{(i,j)_k}$ using a Gaussian basis expansion.

The graph encoder applies $T$ graph convolution layers to iteratively update node representations. At convolution layer $t$, the feature of atom $i$ is updated by aggregating information from its neighboring atoms $j$ connected through edges $(i,j)_k \in \mathcal{E}_i$. For each neighbor interaction, the atom feature $v_i^{(t)}$, neighbor feature $v_j^{(t)}$, and edge feature $u_{(i,j)_k}$ are concatenated to form $z_{(i,j)_k}^{(t)} = v_i^{(t)} \oplus v_j^{(t)} \oplus u_{(i,j)_k}$.
A gated convolution operation is then applied, where a sigmoid function $\sigma(\cdot)$ produces a learned gate and $g(\cdot)$ denotes a nonlinear activation function. The update rule is given by
\begin{align*}
    &v_i^{(t+1)} = v_i^{(t)} \\
    &+ \sum_{(i,j)_k \in \mathcal{E}_i}
\sigma\!\left(z_{(i,j)_k}^{(t)} W_f^{(t)} + b_f^{(t)}\right)
\odot
g\!\left(z_{(i,j)_k}^{(t)} W_s^{(t)} + b_s^{(t)}\right),
\end{align*}
where $W_f^{(t)}$, $W_s^{(t)}$ and $b_f^{(t)}$, $b_s^{(t)}$ are learnable weight matrices and bias vectors at layer $t$, and $\odot$ denotes element-wise multiplication. After $T$ convolution layers, the final node representations $\{v_i^{(T)}\}_{i \in \mathcal{V}}$ encode the local atomic environments. A mean pooling operation over all atoms yields the graph-level representation
\[
\mathcal{Z}_{\mathcal{G}} = \frac{1}{|\mathcal{V}|} \sum_{i \in \mathcal{V}} v_i^{(T)},
\]
where $\mathcal{Z}_{\mathcal{G}} \in \mathbb{R}^{d_{\mathcal{G}}}$ is the crystal graph embedding and serves as a compact structural embedding of the crystal. 

\subsection{Stacking Configuration Representation via AutoEncoder}
\label{subsec:stacking_configuration}

\begin{figure}[ht]
    \centering
    \includegraphics[width=\columnwidth]{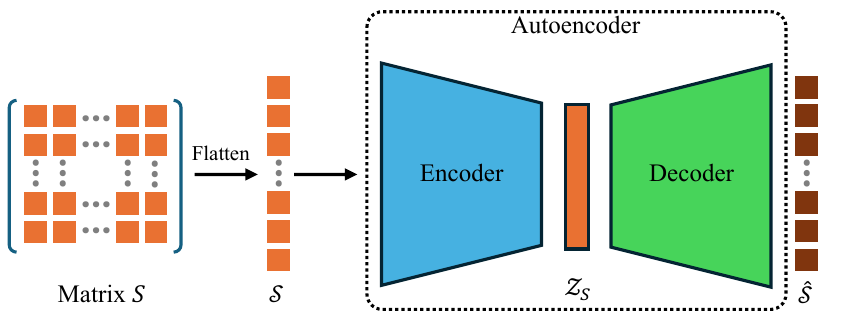}
    \caption{Architecture of extracting stacking configuration representation via Autoencoder ($E_\mathcal{S}$).}
    \label{fig:autoencoder}
\end{figure}

Stacking two-dimensional  materials provides a powerful strategy for engineering novel properties beyond those of the individual layers. By vertically assembling atomically thin monolayers into bilayer heterostructures, we can tune electronic, optical, and mechanical behavior through interlayer coupling, relative orientation, and stacking order. Variations in stacking configuration, such as layer sequence, interlayer distance, and twist angle, can significantly affect the atomic coordinates and the in-plane lattice vectors. As a result, stacked 2D materials may display emergent phenomena not present in isolated monolayers. Stacking configuration is often presented as a matrix and can then be mapped to embeddings. Figure \ref{fig:autoencoder} illustrates the construction of stacking configuration embedding via AutoEncoder.

Algorithm~\ref{alg:stacking_encoder} describes the stacking configuration encoder used to obtain a latent representation of a stacking configuration. Each stacking configuration is represented as a matrix $S \in \mathbb{R}^{n \times m}$ and is flattened into a vector $\mathcal{S} \in \mathbb{R}^{nm}$. The encoder maps the flattened stacking vector $\mathcal{S}$ into a  latent space through a feed-forward network. Specifically, a linear transformation followed by a ReLU activation maps $\mathcal{S}$ to a hidden vector $h_1 \in \mathbb{R}^{d_S/2}$, which is then projected to a latent stacking representation $\mathcal{Z}_S \in \mathbb{R}^{d_S}$. The dimensionality $d_S$ is chosen to be consistent with the graph embedding dimension, enabling fusion with structural graph representations. A symmetric decoder maps the latent vector $\mathcal{Z}_S$ to a hidden vector $h_2 \in \mathbb{R}^{d_S/2}$ using a linear layer followed by a ReLU activation, and then linearly projects $h_2$ to reconstruct the stacking configuration vector $\hat{\mathcal{S}}$. The encoder–decoder network is trained by minimizing the reconstruction loss $\mathcal{L}_{\mathcal{S}} = \lVert \hat{\mathcal{S}} - \mathcal{S} \rVert_2$. After training, the latent vector $\mathcal{Z}_S$ is used as the stacking configuration representation for subsequent multimodal fusion.

\begin{algorithm}[ht]
\caption{Stacking Configuration Encoding ($E_{\mathcal{S}}$)}
\label{alg:stacking_encoder}
\begin{algorithmic}[1]
\setlength{\baselineskip}{1.2\baselineskip}
\Input Stacking configuration matrix $S \in \mathbb{R}^{n \times m}$
\Output Latent stacking representation $\mathcal{Z}_S \in \mathbb{R}^{d_S}$

\State $\mathcal{S} \gets \mathrm{Flatten}(S), \quad \mathcal{S} \in \mathbb{R}^{nm}$

\State $h_1 \gets \mathrm{ReLU}(W_1 \mathcal{S} + b_1), \quad h_1 \in \mathbb{R}^{d_S/2}$
\State $\mathcal{Z}_S \gets W_2 h_1 + b_2, \quad \mathcal{Z}_S \in \mathbb{R}^{d_S}$

\State $h_2 \gets \mathrm{ReLU}(W_3 \mathcal{Z}_S + b_3), \quad h_2 \in \mathbb{R}^{d_S/2}$
\State $\hat{\mathcal{S}} \gets W_4 h_2 + b_4, \quad \hat{\mathcal{S}} \in \mathbb{R}^{nm}$

\State $\mathcal{L}_{\mathcal{S}} \gets \lVert \hat{\mathcal{S}} - \mathcal{S} \rVert_2$
\State Update parameters to minimize $\mathcal{L}_{\mathcal{S}}$

\State \Return $\mathcal{Z}_S$
\end{algorithmic}
\end{algorithm}

\section{Stacking Settings of Bilayer Materials}

Bilayer materials can be broadly classified into homobilayers and heterobilayers based on the composition of their constituent layers. Homobilayer materials are formed by stacking two identical monolayers, where differences in physical properties arise primarily from variations in stacking order, relative translation, or twist angle between the layers. In contrast, heterobilayer materials consist of two distinct monolayers with different chemical compositions or crystal structures, enabling more diverse interlayer interactions and band alignments. 

\subsection{Homobilayer Materials} 
\label{sec:homo-config}

Homogeneous bilayers keep  one monolayer fixed as a reference. Different stacking configurations are obtained by applying a layer-specific transformation of the form $t \circ R \circ F$ to the other monolayer, where $t$ denotes an in-plane translation, $R$ denotes an in-plane rotation, $F$ denotes an optional operation. 
Specifically,  the stacking configuration can be described as a single affine transformation matrix
\[
A =
\begin{bmatrix}
p_1 & p_3 & 0  \\
p_2 & p_4 & 0  \\
0   & 0   & p_5  \\
p_6 & p_7 & \delta z 
\end{bmatrix},
\]
where $p_1$--$p_4$ encode the in-plane rotation $R$ , $p_6$--$p_7$ specify the in-plane translation $t$, and $p_5\in\{+1,-1\}$ represents the optional flip transformation $F$.
$\delta z$ is defined as the fractional shift along the $z$ axis that positions the transformed top monolayer relative to the fixed bottom monolayer
$\delta z = z^{\mathrm{top}} - p_5\, z^{\mathrm{bottom}}$,
where $z^{\mathrm{bottom}}$ and $z^{\mathrm{top}}$ are the fractional $z$ coordinates of corresponding atoms in the bottom and top monolayers, respectively, and $p_5\in\{+1,-1\}$ denotes the flip choice. The value of $\delta z$ is identical for all atom pairs up to numerical precision, serving as a consistency check for the stacking configuration. In general, $\delta z \in [0,1]$ can be chosen by the user to specify the initial distance between two monolayers.

The stacking transformation is applied by multiplying the homogeneous atomic coordinates $(x_i,y_i,z_i,1)$ of each atom in the bottom reference monolayer with the affine transformation matrix $A$, yielding the corresponding atomic coordinates of the top monolayer. Periodic boundary conditions in the in-plane directions are enforced by wrapping the resulting fractional coordinates back into the reference unit cell, which is equivalent to applying a modulo-1 operation. 
Under the row-vector convention adopted in this work, the atomic coordinates of the bottom reference monolayer are collected row-wise in the matrix
$X^{\mathrm{bottom}}$, and the stacking configuration is represented by the affine transformation matrix $A$ with the homogeneous column omitted. The stacking transformation is applied by direct matrix multiplication between the
bottom-layer atomic coordinate matrix and the stacking configuration matrix $X^{\mathrm{top}} = \mathcal{M}(X^{\mathrm{bottom}}A)$, where $\mathcal{M}$ is a function to perform element-wise modulo-by-1 operation. In Appendix \ref{apd:example}, we show an example CIF file and configuration matrix of the bilayer material Al$_4$S$_4$ that is stacked by two monolayer materials of Al$_2$S$_2$.  

For homobilayer materials, since both layers share the same crystal structure, only a monolayer graph is built from its CIF file and then is encoded using the graph encoder $E_{\mathcal{G}}(\cdot)$ to obtain a monolayer-level representation $\mathcal{Z}_{\mathcal{G}} \in \mathbb{R}^{d_{\mathcal{G}}}$. The stacking configuration of the bilayer is independently encoded by the stacking configuration encoder $E_{\mathcal{S}}(\cdot)$ to obtain a stacking representation $\mathcal{Z}_{\mathcal{S}} \in \mathbb{R}^{d_{\mathcal{S}}}$. In addition, monolayer properties are represented by  $\mathcal{P} \in \mathbb{R}^{d_{\mathcal{P}}}$. These representations are concatenated to form a homogeneous bilayer embedding
\begin{equation*}
\mathcal{Z}_{\mathrm{homo}}
=
\mathcal{Z}_{\mathcal{G}}
\oplus
\mathcal{Z}_{\mathcal{S}}
\oplus
\mathcal{P}
\;\in\;
\mathbb{R}^{d_{\mathcal{G}} + d_{\mathcal{S}} + d_{\mathcal{P}}}.
\end{equation*}

\subsection{Heterobilayer Materials} 

A heterobilayer consists of two distinct 2D monolayers stacked via van der Waals interactions. The stacking patterns and sequences in heterostructures generally exhibit greater complexity compared to those in homostructures.  The stacking configuration of heterobilayer is often defined by in-plane relative translation between the two layers, relative rotation (twist angle), interlayer distance, and relative orientation of sublattices and atomic species. 
For hexagonal lattices such as transition-metal dichalcogenides and graphene-based systems, several high-symmetry stacking configurations are commonly considered, including AA stacking, where identical atoms are vertically aligned, and AB or BA stacking, where one layer is shifted so that different atomic species overlap. In heterobilayers, unlike homobilayers, these configurations are generally inequivalent due to broken inversion symmetry, resulting in distinct total energies and electronic structures. The relative stability of different stackings arises from the interplay of electrostatic interactions, interlayer orbital hybridization, and local atomic registry, and while one configuration corresponds to the global energy minimum, metastable stackings may coexist and form experimentally observed stacking domains.

When stacking heterogeneous bilayers,  primitive unit cells of two different monolayers usually cannot be stacked directly due to large lattice mismatch. To resolve this, lattice transformation matrices are applied to each monolayer to construct supercells, and the lattice mismatch between the two supercell monolayers is quantified by the induced in-plane strain. 
For each monolayer pair, multiple supercell pairs are first generated and then filtered by requiring the in-plane strain of both monolayers to remain below a prescribed threshold, together with additional constraints such as a limit on the total number of atoms in the supercell. Among all supercell pairs that satisfy the prescribed criteria, the optimal pair is selected and stacked to form a bilayer structure, which is then subjected to further optimization steps, including interlayer distance optimization and structural relaxation, in order to limit computational cost. 

\section{Experiments}

\subsection{Datasets}

\noindent\textbf{Homobilayer materials.} We use the BiDB dataset, which contains homogeneous bilayers derived from monolayers in the C2DB database. After removing 250 bilayers associated with 10 monolayers lacking CIF files, we obtain 10,899 valid bilayer structures. Eliminating duplicate stacking configurations yields 3,902 unique configurations, which are used to train the autoencoder. We choose bandgap as our target prediction property and exclude samples without bandgap labels, resulting in a final dataset of 6,683 bilayer materials formed from 940 unique monolayers. Since a single monolayer can generate multiple homogeneous bilayers under different stacking configurations, we perform 4-fold cross-validation by splitting the data at the monolayer level rather than the bilayer level. This strategy ensures that bilayers derived from the same monolayer do not appear in both training and test sets, thereby preventing data leakage.
The BiDB stacking descriptor encodes only the in-plane stacking transformation and omits the out-of-plane shift between monolayers. In our experiment , we simply set $\delta z$  as a constant value and use seven parameters $p_1$--$p_7$ of the configuration matrix $A$ discussed in Section \ref{sec:homo-config}.

\noindent\textbf{Heterobilayer materials.} We use the HetDB dataset, which comprises 336 heterogeneous bilayer materials constructed from 38 distinct monolayers. Each bilayer in HetDB is formed by stacking two different monolayers, and no duplicate bilayers are generated from the same monolayer pair under different stacking configurations. Accordingly, we employ 4-fold cross-validation, with data splits performed at the bilayer level, to evaluate model performance on HetDB. We use the twisted angle between the two monolayers provided directly in HetDB as the stacking configuration for each bilayer.
The graph encoder $E_{\mathcal{G}}(\cdot)$ is applied independently to the bottom and top monolayers to obtain $\mathcal{Z}^{{bottom}}_{\mathcal{G}}, \mathcal{Z}^{{top}}_{\mathcal{G}} \in \mathbb{R}^{d_{\mathcal{G}}}$. These embeddings are then combined with the corresponding monolayer property vectors $\mathcal{P}^{{bottom}}, \mathcal{P}^{{top}} \in \mathbb{R}^{d_{\mathcal{P}}}$ and stacking configuration embedding $\mathcal{Z}_{\mathcal{S}} \in \mathbb{R}^{d_{\mathcal{S}}}$ to form the final representation.

\subsection{Experiment Settings}
\label{sec:exp_setting}

\noindent\textbf{BiMat-ML.} In the homobilayer setting, our BiMat-ML fuses the CGCNN-derived monolayer representation, the autoencoder-derived stacking configuration embedding, and the monolayer band gap. The autoencoder is trained for 100 epochs using Adam (learning rate 0.001, batch size 16) to obtain latent stacking representations. The CGCNN model consists of three convolutional layers ($T=3$). 
Atomic neighbors are determined within a cutoff radius $R = 8~\text{\AA}$, with a maximum of $N = 12$ neighbors per atom. The model uses 64-dimensional atomic features and 128 hidden features, and is trained for 500 epochs using stochastic gradient descent (SGD) with a learning rate of 0.001 and a batch size of 128. For heterobilayers, we adopt the same CGCNN architecture and training setup. In HetDB, we additionally include the conduction band minimum (CBM) of both monolayers as input features, since the band gap is defined by the CBM–VBM difference and combining the band gap with either CBM or VBM is sufficient.

\noindent\textbf{Baselines.} 
For comparison, we evaluate several baseline models, including the original CGCNN (referred to as Direct in our experiments) \cite{xie2018cgcnn}, SE-CGCNN \cite{chen2024structural}, SE-MEGNET \cite{chen2024structural} and SE-PAINN \cite{chen2024structural}. For both the BiDB and HetDB datasets, these baselines are trained using the CIFs of stacked bilayer materials and also require access to bilayer CIFs during the inference. Unfortunately, CIFs of stacked bilayers  can only be obtained through costly DFT calculations. 
In contrast, our proposed BiMat-ML does not rely on this strong assumption and instead predicts bilayer properties using only the CIFs and properties of the  monolayers. All experiments are conducted on NVIDIA V100 GPU with 32GB RAM.

\subsection{Experiment Results}
\subsubsection{Performance on Homobilayer Materials}

\begin{table}[ht]
\centering
\renewcommand{\arraystretch}{1.2}
\resizebox{\columnwidth}{!}{
\begin{tabular}{|l|cccc|}
\hline
Model &
MAE $\downarrow$ &
MSE $\downarrow$ &
RMSE $\downarrow$ &
R$^2$ $\uparrow$ \\ \hline

BiMat-ML &
\textbf{0.13}{\scriptsize$\pm$0.02} &
\textbf{0.07}{\scriptsize$\pm$0.02} &
\textbf{0.26}{\scriptsize$\pm$0.05} &
\textbf{0.94}{\scriptsize$\pm$0.02} \\ \hline

{BiMat-ML w/o $\mathcal{P}$} &
{0.35}{\scriptsize$\pm$0.05} &
{0.36}{\scriptsize$\pm$0.11} &
{0.59}{\scriptsize$\pm$0.09} &
{0.68}{\scriptsize$\pm$0.06} \\ \hline

Direct &
0.38 {\scriptsize$\pm$0.02} &
0.38 {\scriptsize$\pm$0.02} &
0.61 {\scriptsize$\pm$0.04} &
0.66 {\scriptsize$\pm$0.02} \\ \hline

SE-CGCNN  &
0.38 {\scriptsize$\pm$0.05} &
0.49 {\scriptsize$\pm$0.19} &
0.69 {\scriptsize$\pm$0.12} &
0.57 {\scriptsize$\pm$0.12} \\ \hline

SE-MEGNET  &
0.36 {\scriptsize$\pm$0.05} &
0.37 {\scriptsize$\pm$0.13} &
0.60 {\scriptsize$\pm$0.11} &
0.67 {\scriptsize$\pm$0.06} \\ \hline

SE-PAINN  &
0.36 {\scriptsize$\pm$0.04} &
0.40 {\scriptsize$\pm$0.08} &
0.63 {\scriptsize$\pm$0.07} &
0.64 {\scriptsize$\pm$0.06} \\ \hline

\end{tabular}
}
\caption{Performance comparison of BiMat-ML and baseline models for bandgap prediction on the BiDB.}
\label{tab:thermo_regression}
\end{table}

Table~\ref{tab:thermo_regression} compares the bandgap prediction performance of BiMat-ML and baseline models on the BiDB dataset. BiMat-ML achieves the best overall performance, exhibiting the lowest MAE (0.13), MSE (0.07), and RMSE (0.26), as well as the highest coefficient of determination ($R^2 = 0.94$). Conventional single-CIF baselines, including Direct, SE-CGCNN and SE-PAINN (all without $\mathcal{P}$), show comparable but inferior performance, with MAE values of around 0.38.

Removing the monolayer property component (BiMat-ML without $\mathcal{P}$) results in a substantial degradation in performance relative to the full BiMat-ML model, with MAE and RMSE increasing to 0.35 and 0.59, respectively, and $R^2$ decreasing to 0.68. Nevertheless, even under this ablation, BiMat-ML outperforms the other baselines across most metrics, including MAE, MSE, and RMSE. These results highlight the critical role of incorporating monolayer property information for accurate bilayer bandgap prediction.

Overall, these results demonstrate that BiMat-ML not only significantly outperforms existing baselines but also benefits strongly from explicitly incorporating monolayer property information, enabling more accurate and robust bilayer bandgap prediction without requiring bilayer CIFs.

\subsubsection{Performance on Heterobilyaer Materials}

\begin{table}[ht]
\centering
\renewcommand{\arraystretch}{1.2}
\resizebox{\columnwidth}{!}{
\begin{tabular}{|l|cccc|}
\hline
 &
MAE $\downarrow$ &
MSE $\downarrow$ &
RMSE $\downarrow$ &
R$^2$ $\uparrow$ \\ \hline
BiMat-ML & 
0.13{\scriptsize$\pm$0.02} & 
0.04{\scriptsize$\pm$0.01} & 
0.21{\scriptsize$\pm$0.03} & 
0.88{\scriptsize$\pm$0.04} \\ \hline

BiMat-ML w/o $\mathcal{P}$ & 
0.16{\scriptsize$\pm$0.01} & 
0.09{\scriptsize$\pm$0.04} & 
0.29{\scriptsize$\pm$0.06} & 
0.76{\scriptsize$\pm$0.12} \\ \hline

BiMat-ML w/o $\mathcal{S}$ & 
0.14{\scriptsize$\pm$0.02} & 
0.05{\scriptsize$\pm$0.02} & 
0.22{\scriptsize$\pm$0.04} & 
0.86{\scriptsize$\pm$0.05} \\ \hline

BiMat-ML w/o $\mathcal{S} \& {P} $ & 
0.16{\scriptsize$\pm$0.02} & 
0.09{\scriptsize$\pm$0.03} & 
0.30{\scriptsize$\pm$0.04} & 
0.73{\scriptsize$\pm$0.09} \\ \hline

Direct  & 
{0.14}{\scriptsize$\pm$0.01} & 
{0.05}{\scriptsize$\pm$0.02} & 
{0.22}{\scriptsize$\pm$0.05} & 
{0.85}{\scriptsize$\pm$0.07} \\ \hline

SE-CGCNN  &
{0.12} {\scriptsize$\pm$0.02} &
{0.03} {\scriptsize$\pm$0.02} &
{0.18} {\scriptsize$\pm$0.02} &
{0.91} {\scriptsize$\pm$0.02} \\ \hline

SE-MEGNET  &
{0.19} {\scriptsize$\pm$0.02} &
{0.09} {\scriptsize$\pm$0.03} &
{0.29} {\scriptsize$\pm$0.06} &
{0.76} {\scriptsize$\pm$0.08} \\ \hline

SE-PAINN   &
\textbf{0.11} {\scriptsize$\pm$0.02} &
\textbf{0.03} {\scriptsize$\pm$0.02} &
\textbf{0.17} {\scriptsize$\pm$0.02} &
\textbf{0.91} {\scriptsize$\pm$0.02} \\ \hline

\end{tabular}
}
\caption{Performance comparison of BiMat-ML and baseline models for bandgap prediction on the HetDB.}
\label{tab:regression_hetdb}
\end{table}

Table~\ref{tab:regression_hetdb} reports the band gap prediction performance of BiMat-ML and baseline models on the HetDB dataset. In contrast to the BiDB results, conventional single-CIF baselines achieve stronger performance on HetDB, with SE-PAINN and SE-CGCNN yielding the lowest errors and highest predictive accuracy. In particular, SE-PAINN achieves the best overall performance, with an MAE of $0.11$, an RMSE of $0.17$, and an $R^2$ of $0.91$, closely followed by SE-CGCNN with comparable accuracy. The original CGCNN model (Direct) also performs competitively, achieving an MAE of $0.14$ and an $R^2$ of $0.85$. BiMat-ML attains strong performance without access to bilayer CIFs, achieving an MAE of $0.13$, an RMSE of $0.21$, and an $R^2$ of $0.88$, outperforming the Direct baseline in terms of MAE and $R^2$. These findings indicate that, for heterogeneous bilayers, direct access to bilayer structures offers a predictive advantage by explicitly capturing complex interlayer interactions and stacking diversity. Nevertheless, BiMat-ML remains competitive without requiring bilayer CIFs derived from computationally expensive DFT calculations, making it a practical and scalable alternative for large-scale bilayer screening.

Ablation results further highlight the contributions of different model components. Removing monolayer property information ($\mathcal{P}$) leads to a noticeable degradation in performance, with the MAE increasing to $0.16$ and the $R^2$ score dropping to $0.76$. In contrast, removing stacking descriptors ($\mathcal{S}$) results in a smaller but consistent decline in accuracy (MAE $=0.14$, $R^2=0.86$). Eliminating both components yields the poorest performance among all BiMat-ML variants, confirming that monolayer properties and stacking information provide complementary predictive value for heterogeneous bilayers. However, compared with the ablation results observed on BiDB, both monolayer property and stacking configuration information are less critical for the HetDB dataset. For HetDB, the correlations between the top and bottom monolayer band gaps and the bilayer band gap are substantially weaker ($r \approx 0.42$ and $0.54$, respectively). In contrast, for BiDB, the monolayer and bilayer band gaps are much more strongly correlated (Pearson $r \approx 0.94$).

\subsubsection{Running Time}

We compare the training and inference runtimes of BiMat-ML and baseline models on both datasets. On BiDB, BiMat-ML and CGCNN-Direct have similar training times (900 s), while SE-PAINN remains comparable (923 s), SE-CGCNN is slower (1501 s), and SE-MEGNET is the slowest (2725 s). On HetDB, BiMat-ML is the most efficient, requiring 100 s for training versus 212 s for CGCNN-Direct, 236 s for SE-CGCNN, 375 s for SE-MEGNET and 110 s for SE-PAINN. All models exhibit identical per-sample inference times (0.5 s). In contrast, density functional theory calculations using VASP \cite{kresse1996efficient} take approximately 4.9 hours on an Intel Xeon Gold 6130H CPU \cite{chen2024structural}. Overall, BiMat-ML achieves competitive or superior efficiency while avoiding costly DFT-based structure calculation.

\section{Conclusion}
In this work, we proposed a multimodal learning framework  that can effectively capture the relationships between the structures and properties of stacked two-dimensional materials.   Our BiMat-ML framework does not require the access of CIFs of stacked bilayer materials as baselines and instead uses two CGCNNs to process CIFs of bottom and top layer materials respectively and effectively fuses them with configuration representation learned by one autoeconder. We emphasize that CIFs of stacked bilayer materials are less available due to very high DFT computational cost.  We evaluated our method with BiDB for  homogeneous bilayers and HetDB for heterogeneous bilayers. Experimental results evaluated with BIDB for homogeneous bilayers and HetDB for heterogeneous bilayers demonstrate effectiveness and efficiency of our framework.  Our BiMat-ML framework is model-agnostic and readily applicable to a range of graph neural network architectures. In future, we will study other GNN architectures such as MEGNET \cite{chen2019graph} and PAINN \cite{schutt2023equivariant}. We also plan to develop algorithms for determining stacking configurations that potentially produce emergent properties in stacked bilayer materials.


\appendix

\section{Example of Stacking Configuration Construction}
\label{apd:example}
\begin{figure}[ht]
    \centering
    \includegraphics[width=1\linewidth]{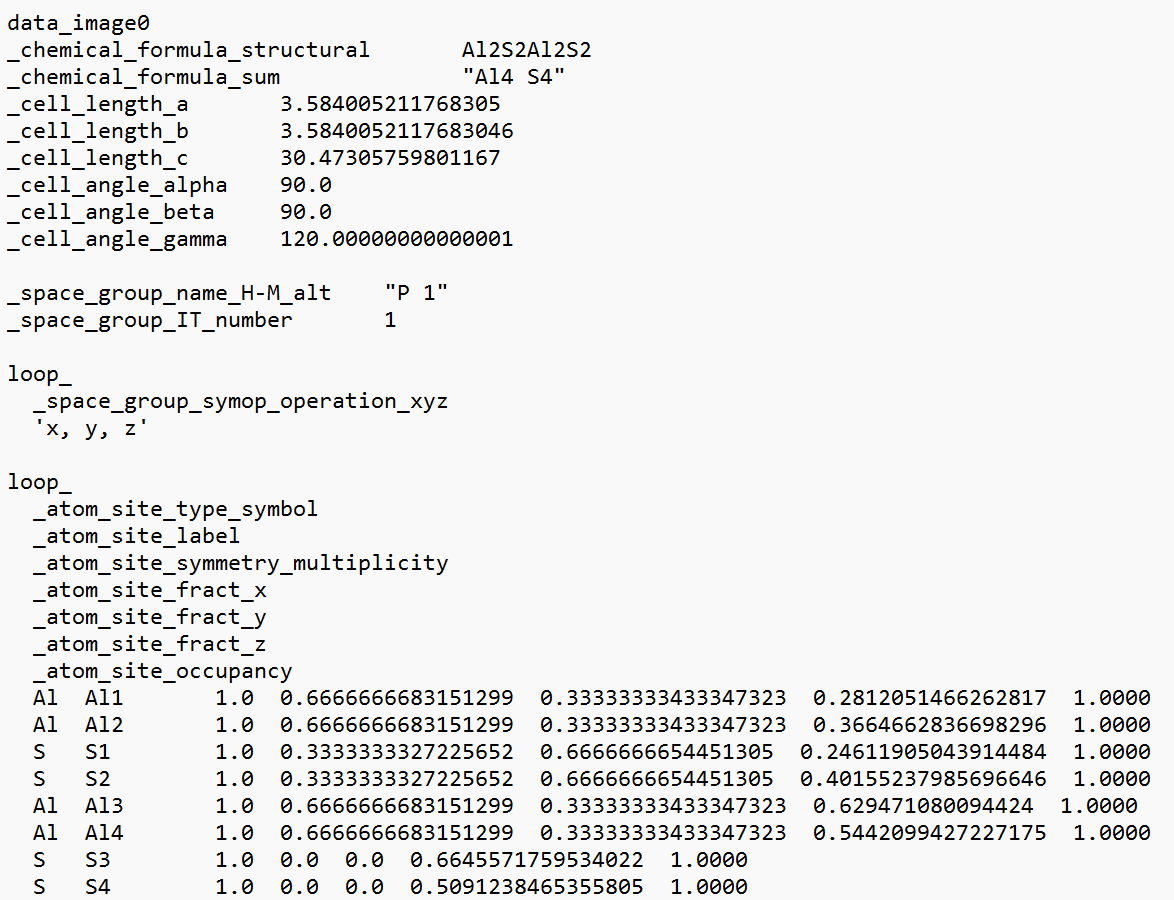}
        \caption{CIF file of an Al$_4$S$_4$ bilayer showing the fractional
        atomic coordinates.}
        \label{fig:delta_z_example}
\end{figure}

We illustrate the stacking configuration construction using the bilayer Al$_4$S$_4$ shown in Figure~\ref{fig:delta_z_example}. The corresponding BiDB stacking descriptor is \texttt{2AlS-2-2--1\_0\_0\_-1-Iz-0.33\_-0.33} which specifies a stacking configuration that includes a flip transformation (\texttt{Iz}, $p_5=-1$). Inspection of the associated CIF file shows that the first four atoms belong to the bottom monolayer, while the remaining four atoms correspond to the top monolayer.

We first determine the out-of-plane shift $\delta z$ from the CIF. For the Al atom pair (Al$_1$, Al$_3$), the fractional $z$ coordinates are $z^{\mathrm{bottom}}=0.28$ and $z^{\mathrm{top}}=0.63$. Using the definition $\delta z = z^{\mathrm{top}} - p_5\, z^{\mathrm{bottom}}$ with $p_5=-1$, we obtain $\delta z = 0.63 + 0.28 = 0.91$. The same value is obtained for all other corresponding atom pairs (e.g., Al$_2$/Al$_4$, S$_1$/S$_3$, S$_2$/S$_4$), confirming that the interlayer shift is uniform across the bilayer. In practice, $\delta z$ can therefore be computed from any corresponding atom pair and used as a consistency check for the stacking configuration.

Next, we construct the stacking transformation from the descriptor parameters. The in-plane rotation parameters are $p_1=-1$, $p_2=0$, $p_3=0$, and $p_4=-1$; the flip parameter is $p_5=-1$; the in-plane translation is $(p_6,p_7)=(0.33,-0.33)$; and the interlayer shift is $\delta z=0.91$. Extracted from the CIF file shown in Figure \ref{fig:delta_z_example}, we have atomic information in  Table \ref{tab:atomic_info}.

\begin{table}[h]
    \centering
    \renewcommand{\arraystretch}{1.2}
    \resizebox{\columnwidth}{!}{
    \begin{tabular}{|l|c|c|c|c|c|c|c|}
        \hline
        Mono. & Type & Label & Multi. & $x$ & $y$ & $z$ & Occu. \\ \hline
        \multirow{4}{*}{Bottom} & Al & Al$_1$ & 1.00 & 0.67 & 0.33 & 0.28 & 1.00  \\
        & Al & Al$_2$ & 1.00 & 0.67 & 0.33 & 0.37 & 1.00 \\
        & S & S$_1$ & 1.00 & 0.33 & 0.67 & 0.25 & 1.00 \\
        & S & S$_2$ & 1.00 & 0.33 & 0.67 & 0.40 & 1.00 \\ \hline
        \multirow{4}{*}{Top} & Al & Al$_3$ & 1.00 & 0.67 & 0.33 & 0.63 & 1.00  \\
        & Al & Al$_4$ & 1.00 & 0.67 & 0.33 & 0.54 & 1.00 \\
        & S & S$_3$ & 1.00 & 0.00 & 0.00 & 0.66 & 1.00 \\
        & S & S$_4$ & 1.00 & 0.00 & 0.00 & 0.51 & 1.00 \\ \hline
    \end{tabular}
    }
    \caption{Atomic information extracted from CIF.}
    \label{tab:atomic_info}
\end{table}

For better mathematical presentation, we re-denote $X^{\mathrm{top}}$ as $X^{\mathrm{t}}$ and $X^{\mathrm{bottom}}$ as $X^{\mathrm{b}}$.
From that, let $X^{\mathrm{b}}$ denote the fractional coordinates of the bottom monolayer atoms in homogeneous coordinates, and let $A$ denote the stacking transformation matrix:

\setlength{\arraycolsep}{2pt}
\begin{align*} 
X_{\mathrm{b}} = \left[ 
\begin{array}{cccc} 0.67 & 0.33 & 0.28 & 1 \\ 
0.67 & 0.33 & 0.37 & 1 \\ 
0.33 & 0.67 & 0.25 & 1 \\ 
0.33 & 0.67 & 0.40 & 1 \end{array} 
\right], 
A = \left[ \begin{array}{ccc} -1 & 0 & 0 \\ 0 & -1 & 0 \\ 0 & 0 & -1 \\ 0.33 & -0.33 & 0.91 \end{array} \right].
\end{align*}

The stacking transformation is applied via matrix multiplication and we can calculate atomic position of top monolayer $X^{\mathrm{t}}$ using inputs from $X^{\mathrm{b}}$ and $A$:
\begin{align*} 
X^{\mathrm{t}} &= \mathcal{M}(X^{\mathrm{b}}\,A) \\
&= \mathcal{M} \left(\left[ \begin{array}{ccc} -0.33 & -0.67 & 0.63 \\ -0.33 & -0.67 & 0.54 \\ 0 & -1.00 & 0.66 \\ 0 & -1.00 & 0.51 \end{array} \right] \right) = \left[ \begin{array}{ccc} 0.67 & 0.33 & 0.63 \\ 0.67 & 0.33 & 0.54 \\ 0.00 & 0.00 & 0.66 \\ 0.00 & 0.00 & 0.51 \end{array} \right] 
\end{align*}
where $\mathcal{M}$ applies a modulo-1 operation to the in-plane coordinates to enforce periodic boundary conditions. This yields the final fractional atomic coordinates of the top monolayer.

\newpage

\section*{Acknowledgements}
This work was supported in part by the National Institute of General Medical Sciences of National Institutes of Health under award P20GM139768, and the Arkansas Integrative Metabolic Research Center at the University of Arkansas.

\bibliographystyle{named}
\bibliography{ijcai26}

\end{document}